\pdfoutput=1
\documentclass[letterpaper, 10 pt, conference]{ieeeconf}  

\IEEEoverridecommandlockouts                              

\usepackage{amsmath} 
\usepackage{amsthm}  
\usepackage{amsfonts}
\usepackage{amssymb}  
\usepackage{dsfont}
\let\labelindent\relax\usepackage{enumitem}
\usepackage[dvipsnames]{xcolor}
\usepackage{tikz}
\usepackage{pgfplots}
\usepackage[hidelinks]{hyperref}
\pgfplotsset{compat=1.18,
	/pgfplots/ybar legend/.style={
		/pgfplots/legend image code/.code={%
			\draw[##1,/tikz/.cd,yshift=-0.25em]
			(0cm,0cm) rectangle (3pt,0.8em);},
}}
\usepackage[nolist,nohyperlinks]{acronym}
\usepackage[capitalize]{cleveref}
\Crefname{figure}{Fig.}{Figs.}
\crefname{equation}{}{}
\Crefname{equation}{Equation}{Equations}

\usepackage[per-mode=symbol,exponent-product=\cdot]{siunitx}

\usepackage{url}
\usepackage[numbers]{natbib}

\usepackage{booktabs}
\usepackage{tabularx}
\newcolumntype{Y}{>{\raggedright\arraybackslash}X}
\newcommand{\otoprule}{\midrule[\heavyrulewidth]}

\let\originalleft\left
\let\originalright\right
\renewcommand{\left}{\mathopen{}\mathclose\bgroup\originalleft}
\renewcommand{\right}{\aftergroup\egroup\originalright}

\newcommand{\fonescore}{\mathrm{F1 score}}
\newcommand{\precision}{\mathrm{Precision}}
\newcommand{\recall}{\mathrm{Recall}}
\newcommand{\result}[1]{R\left(#1\right)}
\newcommand{\scenario}{S}
\newcommand{\scenariopar}{S'}
\newcommand{\scenariocategory}{\mathcal{C}}

\newcommand{\paralonlead}{a_{x,0}^{\mathrm{l}}}
\newcommand{\paralonleadmean}{\bar{a}_{x}^{\mathrm{l}}}
\newcommand{\paralontarget}{a_{x,0}^{\mathrm{c}}}
\newcommand{\pardimension}{d}
\newcommand{\pardistancelead}{D}
\newcommand{\parvlattarget}{v_{y,0}^{\mathrm{c}}}
\newcommand{\parvlonego}{v_{x,0}^{\mathrm{e}}}
\newcommand{\parvlonlead}{v_{x,0}^{\mathrm{l}}}
\newcommand{\parvlonleadfinal}{v_{x,1}^{\mathrm{l}}}
\newcommand{\parvlontarget}{v_{x,0}^{\mathrm{c}}}

\makeatletter
\newcommand*{\org@overidelabel}{}
\let\org@overridelabel\@verridelabel
\renewcommand*{\@verridelabel}[1]{%
	\@bsphack
	\protected@write\@auxout{}{\string\AC@undonewlabel{#1@cref}}%
	\org@overridelabel{#1}%
	\@esphack
}%
\makeatother

\begin{acronym}[AAAAAAAA]
	\acro{ALKS}{Automated Lane Keeping System}\acroindefinite{ALKS}{an}{an}
	\acro{ADS}{Automated Driving System}\acroindefinite{ADS}{an}{an}
	\acro{BTN}{Brake Threat Number}
	\acro{CCHDM}{Careful and Competent Human Driver Model}
	\acro{FN}{False Negative}\acroindefinite{FN}{an}{a}
	\acro{FP}{False Positive}\acroindefinite{FP}{an}{a}
	\acro{FSM}{Fuzzy Safety Model}\acroindefinite{FSM}{an}{a}
	\acro{GAN}{Generative Adversarial Network}
	\acro{LVD}{Leading Vehicle Deceleration}\acroindefinite{LVD}{an}{a}
	\acro{PCA}{Principal Component Analysis}
	\acro{SVD}{Singular Value Decomposition}\acroindefinite{SVD}{an}{a}
	\acro{THW}{Time Headway}
	\acro{TP}{True Positive}
	\acro{TTC}{Time To Collision}
	\acro{R157}[UN~R157]{United Nations Regulation 157}
	\acro{RSS}{Responsibility Sensitive Safety}
\end{acronym}

\newlength{\figurewidth}
\newlength{\figureheight}

\title{\LARGE \bf
Scenario-based assessment of automated driving systems: How (not) to parameterize scenarios?
}

\author{Erwin de Gelder$^{1*}$, Olaf Op den Camp$^{1}$
\thanks{The research presented in this work has been made possible by the SUNRISE project.
	This project is funded by the European Union's Horizon Europe Research \& Innovation Actions under grant agreement No.\ 101069573. 
	Views and opinions expressed are however those of the authors only and do not necessarily reflect those of the European Union or the European Climate, Infrastructure and Environment Executive Agency (CINEA). 
	Neither the European Union nor the granting authority can be held responsible for them.}
\thanks{$^{1}$TNO, Integrated Vehicle Safety, Helmond, The Netherlands}%
\thanks{$^{*}$Corresponding author: {\tt\small erwin.degelder@tno.nl}}%
}

\newcommand{\cstart}{}
\newcommand{\cend}{}

\begin{document}

	\maketitle
	\thispagestyle{empty}
	\pagestyle{empty}

	\begin{abstract}
 
    The development of \acp{ADS} has advanced significantly. 
    To enable their large-scale deployment, the \ac{R157} concerning the approval of \acp{ALKS} has been approved in 2021.
    \Ac{R157} requires an activated \ac{ALKS} to avoid any collisions that are reasonably preventable and proposes a method to distinguish reasonably preventable collisions from unpreventable ones using ``the simulated performance of a skilled and attentive human driver''. 
    With different driver models, benchmarks are set for \acp{ALKS} in three types of scenarios.
    The three types of scenarios considered in the proposed method in \ac{R157} assume a certain parameterization without any further consideration. 
    
    This work investigates the parameterization of these scenarios, showing that the choice of parameterization significantly affects the simulation outcomes.
    By comparing real-world and parameterized scenarios, we show that the influence of parameterization depends on the scenario type, driver model, and evaluation criterion.
    Alternative parameterizations are proposed, leading to results the are closer to the non-parameterized scenarios in terms of recall, precision, and F1 score.
    The study highlights the importance of careful scenario parameterization and suggests improvements to the current UN R157 approach.

\end{abstract}

    \acresetall
	\section{INTRODUCTION}
\label{sec:intro}

\Acp{ADS} are expected to enhance traffic safety by eliminating human errors, providing more comfortable rides, and reduce traffic congestion \citep{chan2017advancements}. 
Lower levels of automation, such as adaptive cruise control \citep{mahdinia2020safety} and lane-keeping assist systems \citep{mammeri2015design}, are already common in modern vehicles. 
Initially, the deployment of higher automation levels (SAE level 3 and above \citep{sae2021j3016}) was hindered by regulations requiring a human driver to be in charge, as per the Vienna Convention on Road Traffic of 1968 \citep{vellinga2019automated}.

The World Forum for Harmonization of Vehicle Regulations approved the \ac{R157} in 2021 for the approval of \acp{ADS}, titled ``Uniform provisions concerning the approval of vehicles with regard to \acp{ALKS}" \citep{ece2021WP29}. 
This is the first regulation considering an automated system that fully takes over the driving task of a human driver for part of the ride.
This regulation states that ``the activated system shall not cause any collisions that are reasonably foreseeable and preventable'' \citep[Clause~5.1.1]{ece2021WP29}. 
A method is proposed in \citep[Annex~3, Appendix~4]{ece2021WP29} to distinguish scenarios with preventable collisions from scenarios with unpreventable collisions ``based on the simulated performance of a skilled and attentive human driver''.
Based on human driver models, benchmarks are set for \acp{ALKS} in three types of scenarios.
The three types of scenarios considered in the proposed method in \ac{R157} assume specific parameterizations without further consideration. 

In this work, we investigate the parameterization of the three types of scenarios in \ac{R157}. 
While parameterizing scenarios enables statistical analyses of the performance of \iac{ADS}, the identification of potential failures of the \ac{ADS}, and the upscaling of scenario-based testing, the particular choice of parameterization appears to have a significant influence on the outcome of the assessment. 
We show this by comparing the simulated performance of the driver models with scenarios as observed in real-world data as well as the same scenarios in a parameterized form.
The results show that the choice of the parameterization has a significant influence on the outcome of the simulated performance that should not be neglected.
We also show that the influence of the parameterization not only depends on the type of scenario, but also on the driver model and the test criteria that are used to evaluate the simulated performance.
For each type of scenario, we propose several alternative parameterizations, and we demonstrate that other choices of scenario parameterization are leading to results that are closer, in terms of recall, precision, and F1 score, to the results with the real-world, non-parameterized scenarios.

This work is structured as follows.
We first explain why and how scenarios are parameterized in \cref{sec:background}.
Our proposed method for analyzing different parameterizations is presented in \cref{sec:method}.
\Cref{sec:setup} explains how we illustrate our method in a case study based on the \ac{R157} scenarios, with results shown in \cref{sec:results}.
\Cref{sec:discussion,sec:conclusions} provide a discussion and conclusions, respectively.

	\section{BACKGROUND}
\label{sec:background}

As mentioned in the introduction, the choice of scenario parameterization significantly impacts assessment outcomes. 
This raises the question: should scenarios be parameterized? 
To address this, we first explain the benefits of parameterization in \cref{sec:why parametrizing}, followed by a review of works on scenario parameterization for assessing \acp{ADS} in \cref{sec:how parametrizing}.

\subsection{Why parameterizing scenarios?}
\label{sec:why parametrizing}

Parameterizing scenarios enables comprehensive testing beyond observed road scenarios. 
Since the number of scenarios \iac{ADS} must be able to handle is virtually infinite, extensive testing with varied scenarios is essential. 
Relying solely on observed scenarios for testing would be impractical and costly as well as making the execution of all test scenarios infeasible.

Another reason to parameterize scenarios is to facilitate statistical analysis. 
By parameterization, we can estimate probability density functions for the parameters, allowing us to assess scenario exposure and quantify the risk associated with \iac{ADS} \citep{degelder2021risk}.

For assessing \iac{ADS}, it is essential to identify the scenarios it might encounter during its lifetime. 
A detailed representation of scenarios, with all state variables defined over time, would result in an impractically large number of scenarios. 
Instead of listing each scenario individually, we can use ranges of valid parameter values to define the scenarios \iac{ADS} must handle. 
The term ``logical scenario,'' introduced in \citep{menzel2018scenarios}, refers to these scenario descriptions with parameter ranges. 
As shown in \citep{nakamura2022defining, degelder2022ss}, using parameter ranges also helps determining the bounds of reasonably foreseeable scenarios.

Due to the vast number of scenarios, assessing \acp{ADS} heavily relies on simulations. 
However, it is impossible to simulate every possible scenario. 
Consequently, research has focused on minimizing the number of simulations by targeting scenarios where the \ac{ADS} exhibits critical behavior. 
A well-known method for this is importance sampling, which automatically selects scenarios for simulation to reduce the number of simulations needed while still providing sufficient confidence in the assessment results, e.g., see \citep{feng2020testing, li2020theoretical, thal2020incorporating}.
This would not be possible if scenarios would not be parameterized.

\subsection{How to parameterize scenarios?}
\label{sec:how parametrizing}

In \ac{R157}, three different types of scenarios are considered: cut-in, cut-out, and \ac{LVD}. 
The parameters of those scenarios are selected without further explanation. 
This approach is also common in many other studies, where specific parameterizations are chosen without additional justification or detailed consideration, e.g., see \citep{feng2020testing, li2020theoretical, thal2020incorporating, nakamura2022defining, degelder2022ss}.

A common challenge with parameterizing scenarios is the so-called ``curse of dimensionality'', where the complexity of estimating statistics and/or performing numerical computations grows exponentially with the number of parameters. 
Therefore, reducing the number of parameters is desirable. 
Several studies focus on techniques for parameter reduction, e.g., see \citep{degelder2021generation, cai2022survey}. 
In \citep{degelder2021generation}, a metric is proposed to find the optimal balance between minimizing the number of parameters for reliable statistics and maximizing parameters to reduce information loss. However, this metric does not account for the impact of parameter reduction on the simulation outcomes of these scenarios, which depends on the sensitivity of the results on the scenario parameters.

Typically, scenario parameters have a physical interpretation. 
However, \acp{GAN} offer a technique that does not rely on such parameters \citep{goodfellow2014generative}. 
\Acp{GAN} have also been applied in the automotive domain \citep{demetriou2020generation, spooner2021generation}. 
It is important to note that \acp{GAN} still use parameters, known as ``latent variables'', which lack physical meaning. 
This essentially makes \acp{GAN} another parameter reduction technique. 
Therefore, with \acp{GAN}, it is still necessary to evaluate whether the automatically determined parameters are suitable.

	\section{METHOD}
\label{sec:method}

Due to the diverse range of scenarios, parameters relevant to one type of scenario may not be applicable to another. 
To differentiate between various types of scenarios, we assume that all scenarios --- which are quantitative descriptions --- can be categorized into one or more scenario categories, with scenarios within the same scenario category being parameterized similarly. 
Here, a scenario category can be seen as the qualitative counterpart of a scenario.
This assumption does not limit the applicability of the proposed methodology, although it may require many scenario categories to cover all scenarios. 
We will denote a scenario category by $\scenariocategory$.

The main idea of the presented method is to compare the simulation outcome of a non-parameterized scenario with the simulation outcome of the corresponding parameterized scenario.
It is important to note that assumptions are also made for the non-parameterized scenario since it is being modeled. 
For instance, a specific sample frequency is chosen.
Additionally, each simulation inherently includes limitations and simplifications of reality.  
When parameterizing a scenario, \emph{additional assumptions} are made to simplify its representation.
The objective of the proposed method is to examine whether the \emph{additional assumptions} used for parameterizing the scenario can be justified.

To formalize the method, let $\result{\scenario} \in \{0, 1\}$ denote the outcome of the simulation of scenario $\scenario$, where $\result{\scenario}=1$ indicates a failure according to a particular criterion and $\result{\scenario}=0$ indicates a pass according to the same criterion. 
For example, $\result{\scenario}=1$ indicates that a collision happened, while $\result{\scenario}=0$ means that the simulation finished without a collision. 
In the case study in \cref{sec:setup}, two additional examples are considered.

To examine the parameterization, a set of non-parameterized scenarios belonging to scenario category $\scenariocategory$ is needed.
For this work, we assume that such a set is available.
One approach to obtain these scenarios is to extract them from real-world data.
A method for identifying scenarios belonging to $\scenariocategory$ is detailed in \citep{degelder2020scenariomining}.
This method involves two steps: First, tags are used to describe activities, such as lane changing
and braking, and statuses, such as ``leading vehicle'' and ``driving slower''. 
Each tag is typically associated with an object and includes a start and end time.
Second, by searching for a specific combination of these tags that define the scenario category $\scenariocategory$, the start and end time of the scenarios can be identified.

To measure the influence of the parameterization on the simulation outcome, we compare $\result{\scenario}$ with $\result{\scenariopar}$, where $\scenariopar$ denotes the parameterized version of $\scenario$.
\Iac{TP} is noted when $\result{\scenario}=\result{\scenariopar}=1$.
If $\result{\scenario}=1$ and $\result{\scenariopar}=0$, \iac{FN} is reported while $\result{\scenario}=0$ and $\result{\scenariopar}=1$ indicates \iac{FP}.
The recall is the ratio of the number of \acp{TP} and the total number of fails when considering the non-parameterized scenarios (\ac{TP}+\ac{FN}) and the precision is the ratio of the number of \acp{TP} and the total number of fails when considering the parameterized scenarios (\ac{TP}+\ac{FP}).
To quantify the influence, we will look at the F1 score, which is the harmonic mean of the recall and the precision:
\begin{equation*}
	\fonescore = 2 \cdot \frac{\recall \cdot \precision}{\recall + \precision}.
\end{equation*}

	\section{SETUP CASE STUDY}
\label{sec:setup}

The scenarios will be discussed first.
Next, we will mention the four different models that are used in this case study in \cref{sec:models}.
\Cref{sec:criteria} lists the three test criteria that are used.
Finally, \cref{sec:parameterizations} presents the different parameterizations.

\subsection{Scenarios}
\label{sec:scenarios}

Following \ac{R157}, three different scenario categories are considered: cut-in, cut-out, and \ac{LVD}.
To obtain the non-parameterized scenarios belonging to these scenario categories, the HighD data set \citep{krajewski2018highD} is chosen.
The data consists of trajectories of cars and trucks at six different locations on German motorways obtained using video footage from drones.

To obtain the scenario data, each of the more than \num{100000} vehicles is treated as an ego vehicle once. I.e., from the total data set, more than \num{100000} smaller data sets are created, where each of the smaller data sets contains a single ego vehicle and trajectory data relative to the ego vehicle as if the other vehicles are perceived from the ego vehicle. 
It is assumed that the ego vehicle can see all of its surrounding vehicles within a distance of \SI{100}{\meter}. 
Each of the smaller data sets stops whenever the ego vehicle is \SI{100}{\meter} from its final position; this is done to avoid the sudden disappearance of vehicles in front of the ego vehicle, as these vehicles would
be out of view of the drone camera. 
In total, this resulted in \num{109986} data sets with a single ego vehicle.

From the data, \num{2992} cut-ins have been found. 
For this study, we consider only those cut-ins where the speed of the vehicle cutting in is less than \SI{95}{\percent} of the ego vehicle's speed and the \ac{THW} is less than \SI{2}{\second}. 
This has resulted in \num{362} cut-ins being used for the experiment.
In total, \num{3069} cut-outs have been identified.
When considering only those cut-outs with another vehicle in front of the vehicle cutting-out that is slower than the ego vehicle, only \num{819} cut-outs were left.
We have found \num{20351} \ac{LVD} scenarios.
To limit the number of \ac{LVD} scenarios for the experiment, only the \num{482} \ac{LVD} scenarios in which the deceleration exceeded \SI{2}{\meter\per\second\squared} have been considered.

\subsection{Models}
\label{sec:models}

To demonstrate that the influence of the parameterization depends on the system-under-test, four different driver reference models are used. 
For the sake of brevity, these models will be summarized shortly; for a more elaborate description, see \citep{mattas2022driver}.

The first model is called Reg157 and is based on paragraph 5.2.5.2 of \ac{R157} that dictates that a collision should be avoided if the \ac{TTC} is above a certain threshold.
Here, the \ac{TTC} is the time remaining until two vehicle collide if they would continue on the same course and speed \citep{hayward1972near}.
With the Reg157 model, when the \ac{TTC} becomes less than a certain threshold (depending on the speed), the ego vehicle decelerates with \SI{6}{\meter\per\second\squared} after a reaction time of \SI{0.35}{\second}.

The second model is the \ac{CCHDM}, which is defined in Appendix~3 of \ac{R157}. 
The ego vehicle with the \ac{CCHDM} brakes with a delay of \SI{0.6}{\second} after the \ac{TTC} is below \SI{2}{\second}.
The \ac{CCHDM} differs from the Reg157 model in that it assumes different stages of braking (releasing foot from acceleration pedal and actual braking, resulting in a deceleration of \SI{0.4}{\meter\per\second\squared} and \SI{7.59}{\meter\per\second\squared}, respectively) and a linear increase of the deceleration with a jerk of \SI{12.65}{\meter\per\second\cubed}.

The third model is based on \citep{shalev2017formal}, which introduces the \ac{RSS} model.
The \ac{RSS} model outlines some constraints that, under specific assumptions, ensure safety if a vehicle adheres to them.
In this case, the ego vehicle will decelerate as soon as both the longitudinal and lateral safety distance margins are violated.

The fourth model is the \ac{FSM} \citep{mattas2020fuzzy}.
Here, the ego vehicle brakes as soon as one of the two safety metrics are nonzero.
The difference with the other three models is that the \ac{FSM} allows for only gentle braking.

\subsection{Pass/fail criteria}
\label{sec:criteria}

The influence of the parameterization depends on the pass/fail criteria that one is interested in. 
To demonstrate this, three different pass/fail criteria are used.
The first criterion is that the ego vehicle should not collide.
The second criterion is that the \ac{TTC}, already mentioned in \cref{sec:models}, should remain above \SI{1}{\second}. 
Note that the \ac{TTC} is only evaluated if the relative lateral position of the vehicle in front does not exceed the width of the ego vehicle.

The third criterion is that the \ac{BTN} should remain below \num{0.8}. 
The \ac{BTN} describes how difficult a collision avoidance maneuver by braking is given a maximum deceleration and jerk. 
If the \ac{BTN} is above \num{1}, it is not possible to avoid a collision, while a \ac{BTN} of \num{0} indicates that there is no threat. 
For the maximum deceleration and jerk, we use the same values as for the \ac{CCHDM}.
Since the original definition of the \ac{BTN} \citep{brannstrom2008situation} requires numerical solving for which convergence cannot be guaranteed up front, we use the modified version presented in \citep{andersson2016multitarget}.
Similar as for the \ac{TTC}, the \ac{BTN} is only evaluated if the relative position of the vehicle in front does not exceed the width of the ego vehicle.

\subsection{Parameterizations}
\label{sec:parameterizations}

As shown in \citep{mattas2022driver}, the initial distance between the ego vehicle and the other vehicle(s) participating in the scenarios has a large influence on the outcome.
If we would use the initial distance as observed in the data, there would be no collision in the simulations as the original data did not contain any (near-)collision.
Therefore, following the approach in \citep{zofka2015datadrivetrafficscenarios}, for all scenarios, the initial \ac{THW} is varied from \SI{2}{\second} down to \SI{0.2}{\second} in steps of \SI{0.2}{\second}.
Consequently, each of the observed scenarios is simulated ten times with varying \ac{THW}.

For the \emph{cut-in} scenarios, the following parameterizations are considered:
\begin{enumerate}
	\item \label{par:cut-in 1} Similar to the parameterization in \ac{R157}, i.e., the initial longitudinal and lateral velocity of the cutting-in vehicle ($\parvlontarget$ and $\parvlattarget$, respectively) and the initial longitudinal velocity of the ego vehicle ($\parvlonego$) are used.
	It is assumed that the longitudinal velocity of the cutting-in vehicle remains constant while the lateral velocity remains constant until the lane change, with a width of \SI{3.5}{\meter}, has been completed.
	
	\item \label{par:cut-in 2} Similar to parameterization~\ref{par:cut-in 1}, but now a constant longitudinal acceleration/deceleration of the cutting-in vehicle ($\paralontarget$) is assumed, equal to the mean acceleration/deceleration in the non-parameterized scenario.
	In case the cutting-in vehicle comes to a standstill, it will remain stationary.
	
	\item \label{par:cut-in 3} This parameterization is based on \citep{degelder2021risk} and is similar to parameterization~\ref{par:cut-in 1}, but now the lane change is assumed to happen instantaneously.
	As a result, $\parvlontarget$ and $\parvlonego$ are the only two parameters.
	
	\item \label{par:cut-in 4} Similar to parameterization~\ref{par:cut-in 2}, but now the lane change is assumed to happen instantaneously.
	
	\item \label{par:cut-in 5} Following the method from \citep{degelder2021generation}, the number of parameters are reduced using \ac{SVD}. 
	The original time series are the lateral position and longitudinal velocity of the cutting-in vehicle and the additional parameters are $\parvlonego$ and the duration of the lane change of the cutting-in vehicle.
	With the parameter reduction, only $\pardimension=3$ parameters are used.
	\cstart Note that \ac{SVD} is employed by \ac{PCA} \citep{abdi2010principal}, so both \ac{SVD} and \ac{PCA} reduce the dimensionality of data while preserving as much variability as possible. \cend
	
	\item \label{par:cut-in 6} Similar to parameterization~\ref{par:cut-in 5}, but with $\pardimension=4$.
	
	\item \label{par:cut-in 7} Similar to parameterization~\ref{par:cut-in 5}, but with $\pardimension=5$.
\end{enumerate}

For the \emph{cut-out} scenarios, the following parameterizations are considered:
\begin{enumerate}[label=\alph*),ref=\alph*]
	\item \label{par:cut-out 1} Mostly similar to the parameterization in \ac{R157}, i.e., the initial longitudinal and lateral velocity of the cutting-out vehicle ($\parvlontarget$ and $\parvlattarget$, respectively) and the initial longitudinal velocity of the ego vehicle ($\parvlonego$) are used.
	It is assumed that the longitudinal velocity of the cutting-out vehicle remains constant while the lateral velocity remains constant until the lane change, with a width of \SI{3.5}{\meter}, has been completed.
	Whereas \ac{R157} considers a stationary vehicle in front of the cutting-out vehicle, we consider a leading vehicle moving with a constant speed ($\parvlonlead$) and an initial distance of $\pardistancelead$ from the cutting-out vehicle.
	
	\item \label{par:cut-out 2} Similar to parameterization~\ref{par:cut-out 1}, but now a constant longitudinal acceleration/deceleration is assumed for the cutting-out vehicle ($\paralontarget$) and the leading vehicle ($\paralonlead$) equal to the respective mean acceleration/deceleration in the non-parameterized scenario. 
	In case any of these vehicles comes to a standstill, it will remain stationary.
	
	\item \label{par:cut-out 3} This parameterization is based on \citep{degelder2021risk} and is similar to parameterization~\ref{par:cut-out 1}, but now the lane change is assumed to happen instantaneously.
	
	\item \label{par:cut-out 4} Similar to parameterization~\ref{par:cut-out 2}, but now the lane change is assumed to happen instantaneously.
	
	\item \label{par:cut-out 5} Following the method from \citep{degelder2021generation}, the number of parameters are reduced using \ac{SVD}.
	The original time series are the lateral position of the cutting-in vehicle and the longitudinal velocity of the cutting-in and leading vehicles. 
	The additional parameters are $\parvlonego$, $\pardistancelead$, and the duration of the lane change.
	With the parameter reduction, only $\pardimension=3$ parameters are used.
	
	\item \label{par:cut-out 6} Similar to parameterization~\ref{par:cut-out 5}, but with $\pardimension=5$.
	
	\item \label{par:cut-out 7} Similar to parameterization~\ref{par:cut-out 5}, but with $\pardimension=7$.
\end{enumerate}

For the \emph{\ac{LVD}} scenarios, the following parameterizations are considered:
\begin{enumerate}[label=\roman*),ref=\roman*]
	\item \label{par:lvd 1} Similar to the parameterization in \ac{R157}, i.e., with the initial longitudinal velocity ($\parvlonlead$), final longitudinal velocity ($\parvlonleadfinal$), and the mean deceleration ($\paralonleadmean$) of the leading vehicle are used.
	It is assumed that the deceleration is linear.
	After the deceleration, the leading vehicle maintains its speed.
	The initial speed of the ego vehicle equals $\parvlonlead$.
	
	\item \label{par:lvd 2} Similar to parameterization~\ref{par:lvd 1}, but now the deceleration follows a sinusoidal shape \citep{degelder2021risk}.
	
	\item \label{par:lvd 3} Using the longitudinal velocity over time of the lead vehicle and the total duration of the deceleration, the number of parameters are reduced following the approach in \citep{degelder2021generation}.
	In total, $\pardimension=3$ parameters are used.
	
	\item \label{par:lvd 4} Similar to parameterization~\ref{par:lvd 3}, but with $\pardimension=4$.
	
	\item \label{par:lvd 5} Similar to parameterization~\ref{par:lvd 3}, but with $\pardimension=5$.
\end{enumerate}

	\section{RESULTS}
\label{sec:results}

\Cref{tab:fails} lists the total number of failures per scenario category, model, and pass/fail criteria for the baseline simulations, i.e., the simulations of the non-parameterized scenarios. 
All models experience a substantial number of collisions in \emph{cut-in scenarios}. 
These collisions include instances where the cutting-in vehicle hits the ego vehicle from the side.
In those cases, the \ac{TTC} and \ac{BTN} are never calculated since the cutting-in vehicle was never in front of the ego vehicle.
This explains why there may be more collisions than simulations where the \ac{TTC} is below \SI{1}{\second}.

\begin{table}
	\centering
	\caption{Number of fails per scenario category, model, and pass/fail criterion for the baseline simulations.
		In total, \num{3620} cut-in, \num{8190} cut-out, and \num{4820} \ac{LVD} simulations are performed.}
	\label{tab:fails}
	\begin{tabular}{llrrr}
    \toprule
    Scenario & Model & Collisions & TTC & BTN \\\otoprule
    Cut-in & Reg157 & \num{349} & \num{2249} & \num{2481} \\
     & CCHDM & \num{538} & \num{457} & \num{677} \\
     & RSS & \num{266} & \num{157} & \num{329} \\
     & FSM & \num{387} & \num{205} & \num{476} \\\otoprule
    Cut-out & Reg157 & \num{195} & \num{4623} & \num{4607} \\
     & CCHDM & \num{65} & \num{1219} & \num{366} \\
     & RSS & \num{21} & \num{94} & \num{65} \\
     & FSM & \num{23} & \num{91} & \num{65} \\\otoprule
    \acs{LVD} & Reg157 & \num{2202} & \num{4819} & \num{4819} \\
     & CCHDM & \num{1303} & \num{3221} & \num{2482} \\
     & RSS & \num{6} & \num{26} & \num{20} \\
     & FSM & \num{15} & \num{32} & \num{22} \\
    \bottomrule
\end{tabular}	
\end{table}

\Cref{tab:fails} shows that the Reg157 model and \ac{CCHDM} frequently fail the \ac{TTC} and \ac{BTN} criteria, with the Reg157 model failing much more often.
This occurs because these models only decelerate when the \ac{TTC} drops below a certain threshold. 
Since both low \ac{TTC} and high \ac{BTN} signals a threat, failing the \ac{TTC} criterion often results in failing the \ac{BTN} criterion.
In contrast, the \ac{RSS} model and \ac{FSM} anticipate earlier to threats, resulting in significantly fewer failures.
Another observation is the high number of collisions for the Reg157 model and \ac{CCHDM} in \emph{\ac{LVD} scenarios}. 
This happens because these models do not account for the leading vehicle's deceleration, causing them to begin braking too late.

\setlength{\figurewidth}{1.1\linewidth}
\setlength{\figureheight}{0.916\figurewidth}
\begin{figure}
	\centering
	\input{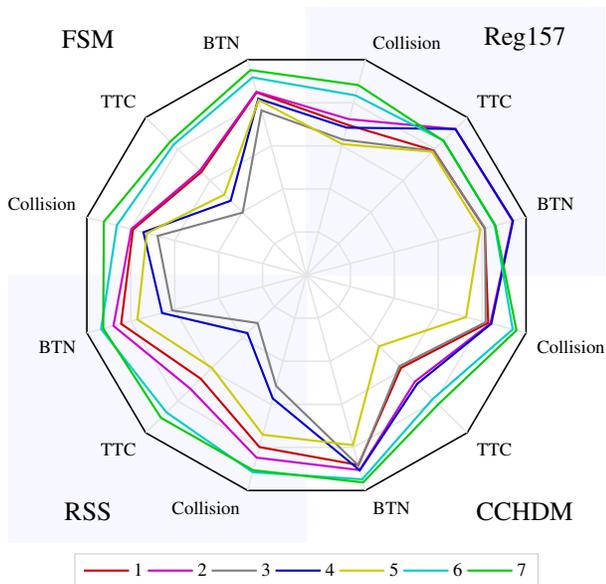}
	\caption{F1 scores for the \emph{cut-in} parameterizations.
		The distance from the center represents the F1 score, starting at \num{0} in the center until a maximum of \num{1} at the outer black line.
		The different lines denote the different parameterizations as listed in \cref{sec:parameterizations}.}
	\label{fig:cut-in results}
\end{figure}

\Cref{fig:cut-in results} shows a radar plot with F1 scores for the \emph{cut-in} parameterizations. 
Adding acceleration as a parameter ($\paralontarget$) improves the F1 scores (parameterization \ref{par:cut-in 2} vs.\ \ref{par:cut-in 1}), though it introduces an extra parameter.
However, assuming an instantaneous lane change by removing parameter $\parvlattarget$ results in significantly worse results, except for the \ac{TTC} and \ac{BTN} criteria with the Reg157 model. 
Generally, assuming an instant lane change leads to many \acp{FP} (meaning that a critical \ac{TTC} or \ac{BTN} is reached in the parameterized scenario but not in the non-parameterized scenario) because the ego vehicle has less time to react in the parameterized scenario.
Consequently, this lowers the precision and F1 score. 
Exceptions are the \ac{TTC} and \ac{BTN} criteria with the Reg157 model, which is highly sensitive to the front vehicle's acceleration, making parameterizations \ref{par:cut-in 2} and \ref{par:cut-in 4} perform best in these cases.
Generally, using the parameter reduction method with $\pardimension=4$ (parameterizations \ref{par:cut-in 6}) or $\pardimension=5$ (parameterizations \ref{par:cut-in 7}) provides the best result in terms of F1 scores.

\begin{figure}
	\centering
	\input{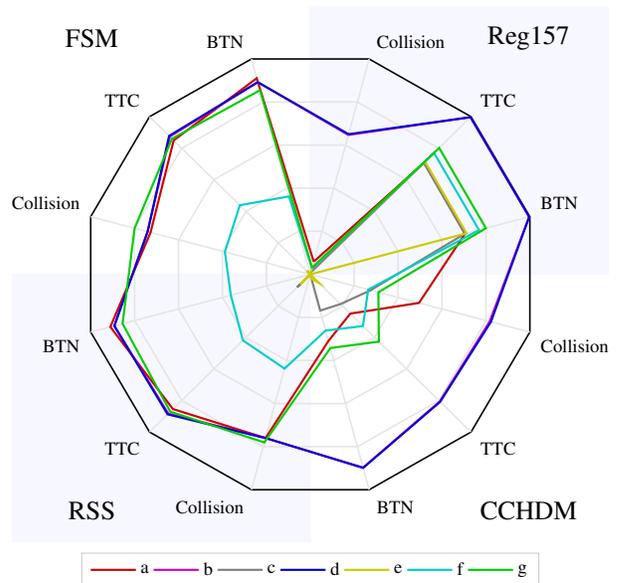}
	\caption{F1 scores for the \emph{cut-out} parameterizations.
		Parameterization \ref{par:cut-out 2} is hardly visible because its results are almost similar to the results of parameterization \ref{par:cut-out 4}.}
	\label{fig:cut-out results}
\end{figure}

For \emph{cut-out} parameterizations, results vary significantly by model and pass/fail criteria, as shown in \cref{fig:cut-out results}. 
With fewer collisions, \acp{FP} or \acp{FN} have a higher impact on the F1 score.
This generally leads to lower F1 scores.
The Reg157 model and \ac{CCHDM} show low F1 scores for all parameterizations except \ref{par:cut-out 2} and \ref{par:cut-out 4} because even slight changes in acceleration/deceleration can alter simulation outcomes.
Parameterizations \ref{par:cut-out 1} and \ref{par:cut-out 3} assume constant speed while parameterizations \ref{par:cut-out 5} to \ref{par:cut-out 7} contain accelerations, but the parameter reduction may cause slight deviations in the actual acceleration/deceleration values.

For the \ac{RSS} model and \ac{FSM}, parameterization \ref{par:cut-out 3} reports an F1 score of almost zero due to the many \acp{FN}, resulting in a low recall.
Similarly, parameterizations \ref{par:cut-out 5} and \ref{par:cut-out 6} have low F1 scores are obtained, suggesting that reducing the parameterization to $\pardimension=3$ or $\pardimension=5$ parameters seems insufficient.

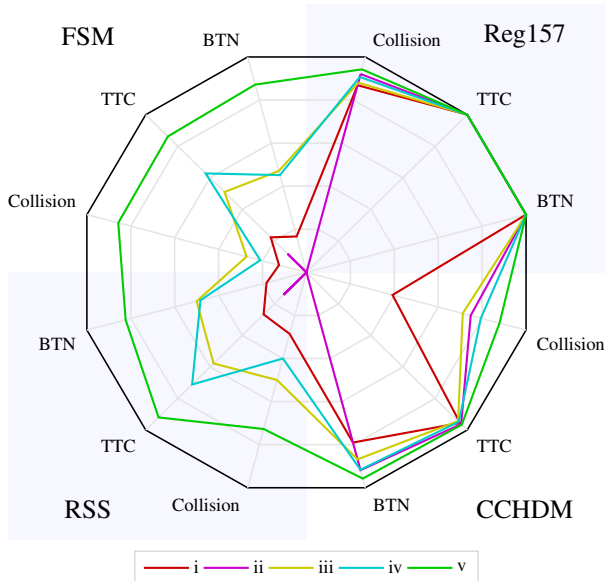
\begin{figure}
	\centering
\begin{tikzpicture}

\definecolor{darkgray176}{RGB}{176,176,176}
\definecolor{darkturquoise0204204}{RGB}{0,204,204}
\definecolor{darkviolet2040204}{RGB}{204,0,204}
\definecolor{gainsboro229}{RGB}{229,229,229}
\definecolor{ghostwhite247247255}{RGB}{247,247,255}
\definecolor{gold2042040}{RGB}{204,204,0}
\definecolor{lightgray204}{RGB}{204,204,204}
\definecolor{lime02040}{RGB}{0,204,0}
\definecolor{red20400}{RGB}{204,0,0}

\begin{axis}[
height=\figureheight,
hide x axis,
hide y axis,
legend cell align={left},
legend columns=5,
legend style={
  fill opacity=0.8,
  draw opacity=1,
  text opacity=1,
  at={(0.5,-0.02)},
  anchor=north,
  draw=lightgray204
},
legend style={nodes={scale=0.7, transform shape}},
scaled y ticks=false,
tick align=outside,
tick pos=left,
width=\figurewidth,
x grid style={darkgray176},
xmin=-1.31, xmax=1.31,
xtick style={color=black},
xticklabel style={align=center},
y grid style={darkgray176},
ymin=-1.2, ymax=1.2,
ytick style={color=black},
yticklabel style={/pgf/number format/fixed,/pgf/number format/precision=3}
]
\path [draw=ghostwhite247247255, fill=ghostwhite247247255]
(axis cs:0,0)
--(axis cs:10,0)
--(axis cs:10,10)
--(axis cs:0,10)
--cycle;
\path [draw=ghostwhite247247255, fill=ghostwhite247247255]
(axis cs:0,0)
--(axis cs:-10,0)
--(axis cs:-10,-10)
--(axis cs:0,-10)
--cycle;
\addplot [semithick, gainsboro229, forget plot]
table {%
0.0517638090205041 0.193185165257814
0.14142135623731 0.14142135623731
0.193185165257814 0.0517638090205042
0.193185165257814 -0.0517638090205041
0.14142135623731 -0.14142135623731
0.0517638090205042 -0.193185165257814
-0.0517638090205042 -0.193185165257814
-0.14142135623731 -0.14142135623731
-0.193185165257814 -0.0517638090205043
-0.193185165257814 0.0517638090205041
-0.14142135623731 0.141421356237309
-0.0517638090205043 0.193185165257814
0.0517638090205041 0.193185165257814
};
\addplot [semithick, gainsboro229, forget plot]
table {%
0.103527618041008 0.386370330515627
0.282842712474619 0.282842712474619
0.386370330515627 0.103527618041008
0.386370330515627 -0.103527618041008
0.282842712474619 -0.282842712474619
0.103527618041008 -0.386370330515627
-0.103527618041008 -0.386370330515627
-0.282842712474619 -0.282842712474619
-0.386370330515627 -0.103527618041009
-0.386370330515627 0.103527618041008
-0.282842712474619 0.282842712474619
-0.103527618041009 0.386370330515627
0.103527618041008 0.386370330515627
};
\addplot [semithick, gainsboro229, forget plot]
table {%
0.155291427061512 0.579555495773441
0.424264068711929 0.424264068711929
0.579555495773441 0.155291427061513
0.579555495773441 -0.155291427061512
0.424264068711929 -0.424264068711928
0.155291427061513 -0.579555495773441
-0.155291427061512 -0.579555495773441
-0.424264068711928 -0.424264068711929
-0.579555495773441 -0.155291427061513
-0.579555495773441 0.155291427061512
-0.424264068711929 0.424264068711928
-0.155291427061513 0.579555495773441
0.155291427061512 0.579555495773441
};
\addplot [semithick, gainsboro229, forget plot]
table {%
0.207055236082017 0.772740661031255
0.565685424949238 0.565685424949238
0.772740661031255 0.207055236082017
0.772740661031255 -0.207055236082017
0.565685424949238 -0.565685424949238
0.207055236082017 -0.772740661031255
-0.207055236082017 -0.772740661031255
-0.565685424949238 -0.565685424949238
-0.772740661031254 -0.207055236082017
-0.772740661031255 0.207055236082016
-0.565685424949239 0.565685424949237
-0.207055236082017 0.772740661031254
0.207055236082016 0.772740661031255
};
\addplot [semithick, black, forget plot]
table {%
0.258819045102521 0.965925826289068
0.707106781186548 0.707106781186548
0.965925826289068 0.258819045102521
0.965925826289068 -0.258819045102521
0.707106781186548 -0.707106781186547
0.258819045102521 -0.965925826289068
-0.258819045102521 -0.965925826289068
-0.707106781186547 -0.707106781186548
-0.965925826289068 -0.258819045102522
-0.965925826289068 0.25881904510252
-0.707106781186548 0.707106781186547
-0.258819045102522 0.965925826289068
0.25881904510252 0.965925826289068
};
\addplot [semithick, gainsboro229, forget plot]
table {%
0 0
0.258819045102521 0.965925826289068
};
\addplot [semithick, gainsboro229, forget plot]
table {%
0 0
0.707106781186548 0.707106781186548
};
\addplot [semithick, gainsboro229, forget plot]
table {%
0 0
0.965925826289068 0.258819045102521
};
\addplot [semithick, gainsboro229, forget plot]
table {%
0 0
0.965925826289068 -0.258819045102521
};
\addplot [semithick, gainsboro229, forget plot]
table {%
0 0
0.707106781186548 -0.707106781186547
};
\addplot [semithick, gainsboro229, forget plot]
table {%
0 0
0.258819045102521 -0.965925826289068
};
\addplot [semithick, gainsboro229, forget plot]
table {%
0 0
-0.258819045102521 -0.965925826289068
};
\addplot [semithick, gainsboro229, forget plot]
table {%
0 0
-0.707106781186547 -0.707106781186548
};
\addplot [semithick, gainsboro229, forget plot]
table {%
0 0
-0.965925826289068 -0.258819045102522
};
\addplot [semithick, gainsboro229, forget plot]
table {%
0 0
-0.965925826289068 0.25881904510252
};
\addplot [semithick, gainsboro229, forget plot]
table {%
0 0
-0.707106781186548 0.707106781186547
};
\addplot [semithick, gainsboro229, forget plot]
table {%
0 0
-0.258819045102522 0.965925826289068
};
\addplot [thick, red20400]
table {%
0.224843635335718 0.839127870831389
0.707106781186548 0.707106781186548
0.965925826289068 0.258819045102521
0.37857611635537 -0.101439164651132
0.67559464519871 -0.67559464519871
0.204508910887567 -0.763237646032975
-0.0739482986007202 -0.275978807511162
-0.188561808316413 -0.188561808316413
-0.175622877507103 -0.0470580082004585
-0.120740728286134 0.032352380637815
-0.157134840263677 0.157134840263677
-0.0431365075170869 0.160987637714845
0.224843635335718 0.83912787083139
};
\addlegendentry{\ref{par:lvd 1}}
\addplot [thick, darkviolet2040204]
table {%
0.238340191544658 0.889497704330363
0.707106781186548 0.707106781186548
0.965925826289068 0.258819045102521
0.72248110584223 -0.19358822885717
0.680492909067225 -0.680492909067225
0.23767636856889 -0.887020283257562
-0 -0
-0.101015254455221 -0.101015254455221
-0 -0
-0 0
-0.0831890330807704 0.0831890330807702
-0 0
0.238340191544658 0.889497704330363
};
\addlegendentry{\ref{par:lvd 2}}
\addplot [thick, gold2042040]
table {%
0.227881166568836 0.850464091722962
0.70703342225085 0.70703342225085
0.965825616119311 0.258792193868461
0.68726821771177 -0.184152963919446
0.667134723313592 -0.667134723313592
0.22465134010735 -0.838410215269067
-0.12940952255126 -0.482962913144534
-0.408550584685561 -0.408550584685561
-0.482962913144534 -0.129409522551261
-0.263434316260655 0.0705870123006874
-0.360485810016672 0.360485810016671
-0.121797197695304 0.454553330018385
0.227881166568835 0.850464091722962
};
\addlegendentry{\ref{par:lvd 3}}
\addplot [thick, darkturquoise0204204]
table {%
0.234768217890732 0.876166917170612
0.70703342225085 0.70703342225085
0.965825616119311 0.258792193868461
0.767081860421587 -0.205538965028527
0.670233114682955 -0.670233114682954
0.237203501969582 -0.885255521083742
-0.103527618041008 -0.386370330515627
-0.502831488843767 -0.502831488843767
-0.466309019587826 -0.124947125221907
-0.203352805534541 0.0544882200215832
-0.443674843097442 0.443674843097441
-0.116886020368881 0.436224566711192
0.234768217890731 0.876166917170612
};
\addlegendentry{\ref{par:lvd 4}}
\addplot [thick, lime02040]
table {%
0.243941725802286 0.910402914780169
0.70703342225085 0.70703342225085
0.965925826289068 0.258819045102521
0.849257138251016 -0.227557764360726
0.68317595363596 -0.68317595363596
0.247829963122835 -0.924914014012339
-0.188232032801833 -0.702491510028413
-0.650538238691624 -0.650538238691624
-0.795468327532174 -0.213145095966782
-0.827936422533487 0.22184489580216
-0.609574811367714 0.609574811367713
-0.225637116243224 0.842089181893034
0.243941725802286 0.910402914780169
};
\addlegendentry{\ref{par:lvd 5}}
\draw (axis cs:0.960660171779821,1.06066017177982) node[
  text=black,
  rotate=0.0
]{Reg157};
\draw (axis cs:0.263995426004571,0.98524434281485) node[
  scale=0.7,
  anchor=south west,
  text=black,
  rotate=0.0
]{Collision};
\draw (axis cs:0.721248916810279,0.721248916810279) node[
  scale=0.7,
  anchor=south west,
  text=black,
  rotate=0.0
]{TTC};
\draw (axis cs:0.98524434281485,0.263995426004571) node[
  scale=0.7,
  anchor=south west,
  text=black,
  rotate=0.0
]{BTN};
\draw (axis cs:0.960660171779821,-1.06066017177982) node[
  text=black,
  rotate=0.0
]{CCHDM};
\draw (axis cs:0.98524434281485,-0.263995426004571) node[
  scale=0.7,
  anchor=north west,
  text=black,
  rotate=0.0
]{Collision};
\draw (axis cs:0.721248916810279,-0.721248916810278) node[
  scale=0.7,
  anchor=north west,
  text=black,
  rotate=0.0
]{TTC};
\draw (axis cs:0.263995426004571,-0.98524434281485) node[
  scale=0.7,
  anchor=north west,
  text=black,
  rotate=0.0
]{BTN};
\draw (axis cs:-0.960660171779821,-1.06066017177982) node[
  text=black,
  rotate=0.0
]{RSS};
\draw (axis cs:-0.263995426004571,-0.98524434281485) node[
  scale=0.7,
  anchor=north east,
  text=black,
  rotate=0.0
]{Collision};
\draw (axis cs:-0.721248916810278,-0.721248916810279) node[
  scale=0.7,
  anchor=north east,
  text=black,
  rotate=0.0
]{TTC};
\draw (axis cs:-0.98524434281485,-0.263995426004572) node[
  scale=0.7,
  anchor=north east,
  text=black,
  rotate=0.0
]{BTN};
\draw (axis cs:-0.960660171779821,1.06066017177982) node[
  text=black,
  rotate=0.0
]{FSM};
\draw (axis cs:-0.98524434281485,0.263995426004571) node[
  scale=0.7,
  anchor=south east,
  text=black,
  rotate=0.0
]{Collision};
\draw (axis cs:-0.721248916810279,0.721248916810278) node[
  scale=0.7,
  anchor=south east,
  text=black,
  rotate=0.0
]{TTC};
\draw (axis cs:-0.263995426004572,0.98524434281485) node[
  scale=0.7,
  anchor=south east,
  text=black,
  rotate=0.0
]{BTN};
\end{axis}

\end{tikzpicture}
	\caption{F1 scores for the \emph{\ac{LVD}} parameterizations.}
	\label{fig:lvd results}
\end{figure}

\Cref{fig:lvd results} displays the F1 scores for the \emph{\ac{LVD} scenarios}. 
All parameterizations achieve perfect F1 scores for the Reg157 model with the TTC and BTN criteria. 
Other than that, however, parameterizations \ref{par:lvd 1} and \ref{par:lvd 2} perform poorly, mainly due to many \acp{FN}.
When parameter reduction is applied (parameterizations \ref{par:lvd 3} to \ref{par:lvd 5}), the F1 scores improve significantly, with parameterization \ref{par:lvd 5} yielding the best results in all cases.

	\section{DISCUSSION}
\label{sec:discussion}

In this work, we have demonstrated that scenario parameterization affects simulation outcomes.
Also, different systems under test and test criteria may require different parameterizations.
Specifically, we have found that the parameterization used in the \ac{R157} might not be optimal.
Although not considered in the current study, the level of detail in a simulator also impacts the optimal parameterization; more detailed simulators likely require more parameters to describe a scenario.
For example, environmental conditions like precipitation could influence systems, and if modeled, would need to be considered as scenario parameter(s).

To measure the impact of parameterizations, we have used the F1 score, which requires both high recall and precision.
Achieving both is generally challenging, so a trade-off is often made to maximize the F1 score.
For safety purposes, focusing solely on recall might be sufficient, which means that \acp{FP} are not penalized. 
This approach could result in a conservative design where a potentially safe system produces failures due to a low precision.

\cstart
In addition to considering metrics like the F1 score, recall, and precision, there are other factors that may influence the choice of parameterization. 
For instance, when estimating a probability density function for the parameters, the number of parameters plays a crucial role in the accuracy of the estimation. 
Generally, the more parameters there are, the lower the accuracy of the estimation tends to be. 
Moreover, having more parameters can make it harder to achieve full coverage of the parameter space when sampling scenario parameter values. 
Thus, even if a parameterization yields a lower F1 score, a simpler model with fewer parameters might sometimes be preferable.
\cend

We have shown that scenario parameterization can significantly influence the simulations outcome, raising the question of whether to parameterize scenarios at all.
In \cref{sec:why parametrizing}, we have explained the benefits of parameterization.
However, combining simulations of both parameterized and non-parameterized scenarios can still be useful.
Replaying real-world, non-parameterized scenarios provides different insights.
In addition, also simulating non-parameterized scenarios helps monitoring if the parameterization is still appropriate.
\cstart Provided that parameterization of scenarios remain useful, future work involves researching the potential of additional parameterization techniques, such as feature selection methods and autoencoders \citep{wang2016auto}. \cend

Four different driver models have been considered in this work, all aiming to represent a ``skilled and attentive human driver'' \citep[Annex~4, Appendix~3]{ece2021WP29}.
As presented in \cref{sec:results}, the four different models displayed significant differences in performance.
This study did not focus on evaluating the actual performance of these models; further research is necessary to establish a true baseline for \iac{ADS}.

    \acresetall
	\section{CONCLUSIONS}
\label{sec:conclusions}

Scenario-based assessment is crucial for evaluating \acp{ADS}. 
The \ac{R157} concerning the approval of \acp{ALKS} recommends a scenario-based approach to benchmark for \acp{ALKS} against ``the simulated performance of a skilled and attentive human driver''.
In this study, we have examined the scenario parameterizations proposed in \ac{R157} and demonstrated that they significantly impact simulation outcomes.
This paper shows the need for careful consideration when adopting specific scenario parameterizations and that the optimal choice of a parameterization depends on factors like the system under test and the test criteria.
We have proposed a method to assess the scenario parameterization and, by applying it to the \ac{R157} scenarios, identified and tested alternatives parameterizations that show potential improvements over the scenario parameterization currently used in \ac{R157}. 

\cstart Based on the research presented in this work, we recommend that future amendments to \ac{R157} include an additional requirement. 
Currently, the system's performance in parameterized scenarios is compared with the performance of a human reference model to show that the activated system does not cause any collisions that are reasonably preventable. 
In addition, it should be necessary to justify that the chosen parameterization of scenarios is appropriate. 
This justification should consider the system's performance, the type of scenario, and the specific metrics of interest. \cend

	\addtolength{\textheight}{-5cm}  

	{\footnotesize\bibliographystyle{abbrvnat}
	\bibliography{bib}}

\end{document}